\documentclass[abstracton]{scrartcl} 
\usepackage{epsf, pstricks, graphics, graphicx} 
\usepackage{multirow} 
\usepackage{endnotes} 
\usepackage[backend=biber, citestyle=apa, bibstyle=apa, sortcites]{biblatex}
\usepackage{float} 
\usepackage{marvosym} 
\usepackage{threeparttable}
\usepackage{rotating} 
\usepackage{setspace} 
\usepackage{multicol} 
\usepackage[T1]{fontenc} 
\usepackage[utf8]{inputenc}
\usepackage{amsmath}
\usepackage{amssymb}
\usepackage{mathcomp}
\usepackage{lscape}
\usepackage{multicol}
\usepackage{framed}
\usepackage[german,english]{babel}
\newcommand{\x}[1]{\textit{#1}}

\begin{document} 
\title{Why machines do not understand: A response to S{\o}gaard}
\author{Jobst Landgrebe \and Barry Smith}
\date{\today}
\publishers{State University of New York at Buffao, Department of Philosophy, \texttt{[jobstlan|phismith]@buffalo.edu}}
\selectlanguage{english}
\maketitle

\begin{abstract}
Some defenders of so-called `artificial intelligence' believe that machines can understand language. In particular, S{\o}gaard has argued in this journal for a thesis of this sort, on the basis of the idea (1) that where there is semantics there is also understanding and (2) that machines are not only capable of what he calls `inferential semantics', but even that they can (with the help of inputs from sensors) `learn' referential semantics \parencite{sogaard:2022}. We show that he goes wrong because he pays insufficient attention to the difference between language as used by humans and the sequences of inert of symbols which arise when language is stored on hard drives or in books in libraries. 
\end{abstract}

\section{Large Language Models (LLMs) have no semantics and no pragmatics}

\subsection{How LLMs are created and what they do}

So-called large language models (LLMs), such as the ones built into chatGPT and GPT-4, contain encodings of natural language symbol sequences which represent morphological and syntactic relationships between their constituent symbols. This means that a model of this sort can represent both the internal structure of words and the ways in which words are put together to form phrases, sentences and paragraphs. There is nothing here which the machine could be said to \x{understand} \parencite{landgrebesmith:2021, landgrebesmith:2022}. The machine works, after all, by applying a gigantically long equation (an operator in the language of functional analysis) to generate new outputs from its inputs. The operator's  parameters are determined by the linguistic sample data with which it was, as they say, `trained'. 

The coding of the positions of symbols in symbol sequences in the LLM is enabled through use of the transformer deep neural network (dNN) architecture \parencite{vaswani:2017}. This allows the model to relate positions within a sequence, for example the distance between symbols within a larger text in order to compute a representation of the sequence in a way which takes into account its structure. The models can also incorporate parameters which encode lexicographic similarity relations such as synonymy, and simple lexematic relations such as family relationships or semantic fields such as tree species. The scope of these relationships depends on the structure and size of the models and has been expanded significantly since the transformer dNN architecture was first published in 2017.

The models are created by taking samples from existing corpora, and attempting to create an operator that can be parameterised to \x{fit} these samples. This parameterisation process is erroneously called `training', but it is in fact achieved by using non-linear programming to minimise a loss function in a way which yields a local optimum as in any multivariate optimisation problem (\parencite{bertsekas:2016}, \parencite{goodfellow:2016}).\footnote{The results are then what we can think of as `implicit' mathematics, as contrasted with the usual (explicit) mathematics which involves the construction by the mathematician of operators, functions, and so forth, through acts of thinking. Implicit mathematics is the creation of mathematical equations through a process that involves feeding externally derived data into a machine. The results of this process may involve billions or trillions of parameters.} 
This leads to a stochastic model of the multidimensional distribution of symbols in the sample corpus (`training material'). The resulting stochastic distribution is then an approximative mathematical model of the sample corpus; given any symbol in a string in that corpus it computes the probability of the next most likely occurring string.

Based on this distribution model, an LLM can be used to build very impressive language applications relating to the entire corpus from which samples were drawn by creating sequence outputs based on conditioned probabilities. That is, the model can complement a sequence given as input by extending it with one further symbol, and then extend the result with one further symbol, and so on, in the now familiar ways. Conditioned probabilities can in this way be used to answer questions or fulfil tasks. If the question or task is similar to the symbol sequence samples used for the training of the model, the yielded output can be very convincing. If, however, a model is given an input sequence that is not represented or is underrepresented in the data they were trained with, then its performance declines to the extent that it will generate so-called `hallucinations', which are sequences of symbols that do not correspond to reality, such as incorrect birth and death dates of a person, newly invented journal articles, or entire valleys in Switzerland. 

\subsection{Why LLMs have neither semantics nor pragmatics}

It is regularly claimed, as by \textcite{sogaard:2022}, that such models can mathematically map sentence semantics or even text pragmatics. The ability of LLMs to be used for classification or translation is often mentioned as an example of such semantic capabilities, though semantics are in fact needed for neither: Classification and conditioned sequence-regression are traditional tasks of statistical learning using the above described loss function, and they have nothing at all to do with semantics.

What then are the core arguments in favour of postulating LMM semantics? 

In traditional philosophical usage, there are two types of semantics: 1. that which is concerned with the intentions of people in making utterances (what they \x{mean}), specifically in connection with what they are referring to when they use language, and with associated questions of truth and falsehood;  and 2. that which is concerned with systems of axioms, or with systems of logically formulated propositions more generally, typically involving the creation of set-theoretic models. The latter can be implemented in Turing machines using mechanical theorem provers, which are decidable for sets of propositional logic axioms and semi-decidable for axioms using first-order logic. No one, to our knowledge, has suggested that theorem provers thereby `understand' the formulae to which they are applied.

S{\o}gaard himself does not however refer to this second type of semantics. Rather, the distinction he wants to draw -- documented in \parencite{marconi:1997} -- is between  `inferential' and `referential' semantics. The former is described as `the part of semantics that is concerned with valid inference' and thus (S{\o}gaard believes) it is about relationships between words and sentences. 

But given that dNNs can model relationships between symbols in a morpho-syntactic sense both at and above the sentence level, calling this ability `inferential \x{semantics}' creates only the illusion that these models are indeed capable of semantic modelling. 

`Referential semantics' on the other hand, is said to be about `truth conditions, mental representations or situations in which using a word is deemed appropriate'. 

In the end, humans do indeed use referential semantics in something like the sense defined when they relate reality and the language representing reality on the basis of their intentions, whether as utterer or as interpreter. 

But S{\o}gaard's distinction between inferential and referential semantics is void. For as concerns humans, the relationships between words which form the subject matter of inferential semantics result from the human usage of language about reality, and thus from referential semantics. But as concerns machines, inferential semantics is just syntax, and talk of referential semantics reflects simply an anthromorphic projection onto the machine of these same human relations to reality.

We perform speech acts of uttering, and we engage in internal acts of language use, for example in interpreting someone else's utterance when engaging in a conversation. We do this because we have intentions. These sentences contain the relationships they do because we intend them to do so. But once a sentence is recorded or written down, the immediate connection to the utterer's intentions is lost. Within the sentence itself, there remains only a trace of these intentions in the form of certain syntactic connections. And when a model gets parameterised using a large language corpus, it is these syntactic connections which are modelled by the resulting multivariate stochastic distribution. The latter has nothing to do with inference, which is an exclusively human capability.

Something similar is true for text pragmatics. When S{\o}gaard says that his inferential semantics is about relationships between sentences, he is actually referring to inferential (text) \x{pragmatics}, for it is pragmatics which deals with multi-sentence utterances \parencite{verschueren:1999}. 

Viewed through the single-sentence lens of semantics, when multi-sentence utterances are written down, that aspect of their meaning which requires intentionality is lost. Only the relationships between the symbols, so only syntax, remains.

Only human readers can recover this meaning by the act of interpretation. The best that machines can do is to identify patterns that recur in similar texts and thereby engage in implicit modelling of synonymy or recurring lexematic fields such as family relationships or military hierarchies. S{\o}gaard sees such semantic fields as `inferential semantics' inside the machine. In reality, however, these fields exist where this is human intentionality as expressed in acts of speaking, understanding and thinking. Machines have no knowledge of the semantisc fields, they do not represent or manipulate them as humans do. When machines get parameterised, such lexical relationships are not learned through a drawing of inferences; rather they become part of the multivariate distribution at the level of meaningless symbols as a result of mathematical operations applied to binary vectors. The machine does not understand synonymy, even if it can to a certain extent model associated patterns on the level of syntax.

The ultimate form of what S{\o}gaard  calls `inferential semantics' would consist in the machine having `knowledge' of the reality which its outputs, when interpreted by humans, can be seen as describing. We can view this `knowledge' as it exists in the human context as a collection of propositions about reality; as it exists in the machine, however, it is a structure or pattern that has floated free from its human creators, and thereby shorn of meaning. `Interpretation' of this `knowledge' means the reconstitution (reanimation) of the meanings of the corresponding sentences by a human interpreter. Machines do not do anything like this. LLMs and other stochastic AI models merely compute (conditioned) probabilities over large multivariate distributions encoded by the model parameters. 

The idea that we can describe machines as `learning' or `understanding' because they use `inferential semantics' reflects merely the familiar pattern in AI literature of using \x{Ersatz}-definitions (as when someone defines `flying' as ‘moving in the air’, and then jumping up and down while shouting: ‘See, I am flying!’ \parencite[p. 59]{landgrebesmith:2022}. What linguists and philosophers of language call syntax or lexemic structures is renamed as `inferential semantics' in order to be able to attribute semantics to machines. What results does not reflect in any way the mathematical processes which machines perform, and nor does it challenge Searle's key insight 
that machines perform merely syntactic operations which are ontologically objective and observer-independent \parencite{searle:1980}. The processes inside the machine have no meaning for a non-human observer; the attribution of meaning to them is observer dependent. And because consciousness is observer independent, we can never realistically assign the properties of consciousness to the processes inside a machine \parencite[ch.~9]{landgrebesmith:2022}. 

\section{Machines will never understand anything}

But couldn't we find a way to make machines understand, somehow, in the future, and in some literal sense of the word `understand'? A huge body of literature in favour of this idea rests on the argument that machines perform operations similar to the operations of the human mind. And since the latter are the basis of consciousness, will, intelligence, and understanding, then there surely is no way to rule out the achievement of the latter on the part of the machine. S{\o}gaard, too, proposes a thesis along these lines (loc. cit., p. 5). If the mind was just a machine, we could indeed emulate it with a computer; it would just be a matter of time until we succeed in doing so. In our recent monograph \parencite{landgrebesmith:2022}, however, we show that, because machines are logic systems, they can at best create only partial emulations of complex systems like the human mind-body-continuum (in the same way a weather forecast is a partial emulation of the actual weather).

S{\o}gaard tries to give an example of a radio-camera-machine that could `learn' referential semantics. He supposes that this machine has a will (and therefore a consciousness) which motivates it to understand the world. Our book shows also, however, that machines will never have a will, or consciousness, because these capabilities, too, are impossible to model mathematically. And as Turing pointed out, only what can be modelled mathematically can be emulated in a machine \parencite{turing:1936}.

\printbibliography
\end{document}